\pdfoutput=1

\documentclass[11pt]{article}

\usepackage[]{naacl2021}

\usepackage{times}
\usepackage{latexsym}

\usepackage[T1]{fontenc}

\usepackage[utf8]{inputenc}

\usepackage{microtype}
\usepackage{parskip}
\usepackage{microtype}
\usepackage{bm}
\usepackage{subcaption}
\usepackage{amsmath}
\usepackage{xcolor}
\usepackage{cleveref}
\usepackage{todonotes}
\usepackage{multirow}
\usepackage{booktabs}
\usepackage{xpatch}
\usepackage{blindtext}
\usepackage{algorithm}
\usepackage{algpseudocode}

\makeatletter
\xpatchcmd{\paragraph}{.05ex \@plus1ex \@minus.1ex}{0.05pt plus 0.1pt minus 0.1pt}{\typeout{success!}}{\typeout{failure!}}
\makeatother

\newcommand{\DYGIEPP}[1]{\textsc{DyGIE++}}
\newcommand{\BASE}[1]{\textsc{Base}}
\newcommand{\DEED}[1]{\textsc{DeeD}}
\newcommand{\DCRF}[1]{\textsc{BCrf}}
\newcommand{\DVN}[1]{\textsc{DVN}}
\newcommand{\DocTrigger}[1]{\textsc{DocTrigger}}
\newcommand{\DocArgument}[1]{\textsc{DocArgument}}
\newcommand{\DocEvent}[1]{\textsc{DocEvent}}
\newcommand{\OneIE}[1]{$\text{OneIE}^\text{+}$}

\title{Document-level Event Extraction with Efficient End-to-end Learning of Cross-event Dependencies}

\author{
     Kung-Hsiang Huang\textsuperscript{\rm 1} ~~
     Nanyun Peng\textsuperscript{\rm 1,2} \\
     \textsuperscript{\rm 1} Information Sciences Institute, University of Southern California\\
    \textsuperscript{\rm 2} Computer Science Department, University of California, Los Angeles \\
    {\tt kunghsia@usc.edu,} 
    {\tt violetpeng@cs.ucla.edu} 
}

\begin{document}
\maketitle

\begin{abstract}
Fully understanding narratives often requires identifying events in the context of whole documents and modeling the event relations. However, document-level event extraction is a challenging task as it requires the extraction of event and entity coreference, and capturing arguments that span across different sentences. 
Existing works on event extraction usually confine on extracting events from single sentences, which fail to capture the relationships between the event mentions at the scale of a document, as well as the event arguments that appear in a different sentence than the event trigger. In this paper, we propose an end-to-end model leveraging Deep Value Networks (\DVN~), a structured prediction algorithm, to efficiently capture cross-event dependencies for document-level event extraction. Experimental results show that our approach achieves comparable performance to CRF-based models on ACE05, while enjoys significantly higher computational efficiency.
\end{abstract}

\section{Introduction}

Narratives are account of a series of related events or experiences \cite{Laurence:68}. Extracting events in literature can help machines better understand the underlying narratives. A robust event extraction system is therefore crucial for fully understanding narratives.


\begin{figure}[t]
    \centering
    \begin{subfigure}{0.48\textwidth}

    \includegraphics[width=.95\linewidth]{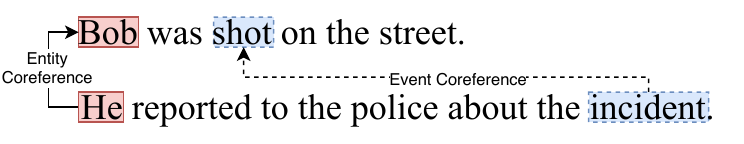}
    \caption{Coreference Example.}
    \label{fig:coref_toy}
    \end{subfigure}
    
    \begin{subfigure}[b]{0.48\textwidth}
    \includegraphics[width=.95\linewidth]{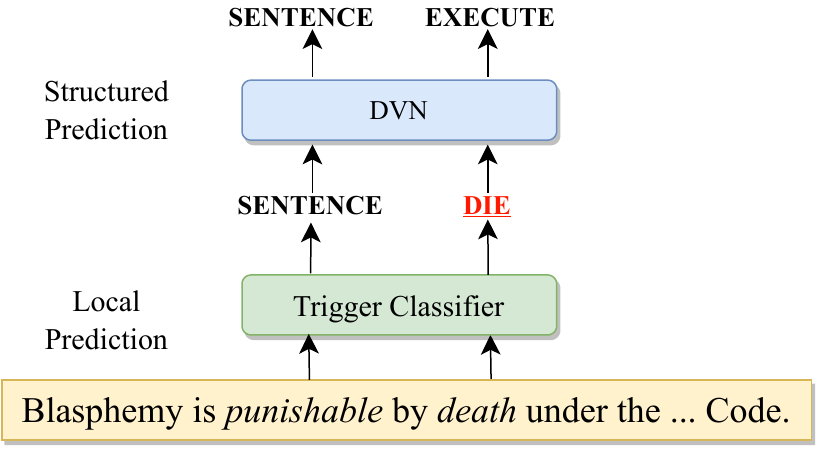}
    \caption{Cross-event Dependencies Example.}
    \label{fig:dvn_toy}
    \end{subfigure}
    
    \caption{ (a) demonstrates why coreference resolution is essential for event extraction. In the second sentence, without entity coreference, an event extraction system cannot identify which real-world entity does \textit{He} refer to. Similarly, \textit{incidence} and \textit{shot} will be incorrectly linked to two different real-world events without event coreference. (b) shows the importance of cross-event dependencies. The local trigger classifier falsely classifies \textit{death} as type \textsc{Die}. Instead, it is an \textsc{Execute} event as a person's life is taken away by an authority. A structured prediction model that learns cross-event interactions can potentially infer the correct event type for \textit{death} given the previous \textsc{Sentence} event is often carried out by authorities.}
    \vspace{-5mm}
\end{figure}

Event extraction aims to identify events composed of a trigger of pre-defined types and the corresponding arguments from plain text \cite{Grishman2005NYUsEA}. To gain full information about the extracted events, entity coreference and event coreference are important, as demonstrated in \Cref{fig:coref_toy}. These two tasks require document-level modeling. The majority of the previous event extraction works focus on sentence level \cite{li-ji-2014-incremental, huang-etal-2020-biomedical, lin-etal-2020-joint}. Some later works leverage document-level features, but still extract events at the scope of sentence \cite{yang-mitchell-2016-joint,zhao-etal-2018-document, Wadden2019EntityRA}. More recently, \citet{du-cardie-2020-document} and \citet{du2020document} treat document-level event extraction as a template-filling task. \citet{li-etal-2020-gaia} performs event mention extraction and the two coreference tasks independently using a pipeline approach. However, none of the previous works learn entity and event coreference jointly with event mention extraction. We hypothesize that joint learning event mention extraction, event coreference, and entity coreference can result in richer representations and better performance.

Moreover, learning cross-event dependencies is crucial for event extraction. \Cref{fig:dvn_toy} shows a real example from the ACE05 dataset on how learning dependencies among event mentions can help correct errors made by local trigger classifiers. However, efficiency is a challenge when modeling such dependencies at the scale of document. While some works attempted to capture such dependencies with conditional random field or other structured prediction algorithms on hand-crafted features \cite{li-etal-2013-joint,lin-etal-2020-joint}, these approaches subject to scalablility issue and require certain level of human efforts. In this work, we study end-to-end learning methods of an efficient energy-based structured prediction algorithm, Deep Value Networks (\DVN~), for document-level event extraction. 

The contribution of this work is two-fold. First, we propose a document-level event extraction model, \DEED~ (\textbf{D}ocument-level \textbf{E}vent \textbf{E}xtraction with \textbf{D}VN). \DEED~ utilizes \DVN~ for capturing cross-event dependencies while simultaneously handling event mention extraction, event coreference, and entity coreference. Using gradient ascent to produce structured trigger prediction, \DEED~ enjoys a significant advantage on efficienty for capturing inter-event dependencies. Second, to accommodate evaluation at the document level, we propose two evaluation metrics for document-level event extraction. Experimental results show that the proposed approach achieve comparable performance with much better training and inference efficiency than strong baselines on the ACE05 dataset. 


\label{Sec:Intro}

\section{Related Works}
In this section, we summarize existing works on document-level information extraction and event extraction, and the application of structured prediction to event extraction tasks.

\paragraph{Document-level Information Extraction} 
Information extraction (IE) is mostly studied at the scope of sentence by early works.~\cite{ju-etal-2018-neural, qin-etal-2018-robust, stanovsky-etal-2018-supervised}. Recently, there has been increasing interest in extracting information at the document-level. \citet{jia-etal-2019-document} proposed a multiscale mechanism that aggregates mention-level representations into entity-level representations for document-level \textit{N}-ary relation extraction. \citet{jain-etal-2020-scirex} presented a dataset for salient entity identification and document-level \textit{N}-ary relation extraction in scientific domain. \citet{li2020exploiting} utilized a sequence labeling model with feature extractors at different level for document-level relation extraction in biomedical domain. \citet{hu2020leveraging} leveraged contextual information of multi-token entities for document-level named entity recognition. A few studies which tackled document-level event extraction will be reviewed in \Cref{subsec:related-work_docEE}.

\paragraph{Document-level Event Extraction} \label{subsec:related-work_docEE}
Similar to other IE tasks, most event extraction methods make predictions within sentences. Initial attempts on event extraction relied on hand-crafted features and a pipeline architecture~\cite{ahn-2006-stages, gupta-ji-2009-predicting, li-etal-2013-joint}. Later studies gained significant improvement from neural approaches, especially large pre-trained language models~\cite{Wadden2019EntityRA, nguyen-etal-2016-joint-event, Liu2018JointlyME, lin-etal-2020-joint, BALALI2020106492}. Recently, event extraction at the document level gains more attention. \citet{yang-etal-2018-dcfee} proposed a two-stage framework for Chinese financial event extraction: 1) sentence-level sequence tagging, and 2) document-level key event detection and heuristic-based argument completion. \citet{zheng-etal-2019-doc2edag} transforms tabular event data into entity-based directed acyclic graphs to tackle the \textit{argument scattering} challenge. \citet{du-cardie-2020-document} employed a mutli-granularity reader to aggregate representations from different levels of granularity. However, none of these approaches handle entity coreference and event coreference jointly. Our work focus on extracting events at the scope of document, while jointly resolving both event and entity coreference.

\paragraph{Structured Prediction on Event Extraction}
Existing event extraction systems integrating structured prediction typically uses conditional random fields (CRFs) to capture dependencies between predicted events~\cite{xu2019jointly, wang2018bidirectional}. However, CRF is only applicable to modeling linear dependencies, and has scalablility issue as the computation cost at least grows quadratically in the size of label. Another line of solutions incorporated beam search with structured prediction algorithms.
\citet{li-etal-2013-joint} leveraged structured perceptron to learn from hand-crafted global features. \citet{lin-etal-2020-joint} adopted hand-crafted global features with a global scoring function and uses beam search for inference. While these structured prediction methods can model beyond linear dependencies and alleviate the scalability issue, it requires pre-defined orders for running beam search. In contrast, our method addresses the above two issues by adopting an efficient stuctured prediction algorithm, Deep Value Networks, which runs linear in the size of label and does not require pre-defined order for decoding.

\section{Document-level Event Extraction}
\subsection{Task Definition}
The input to the document-level event extraction task is a document of tokens $\mathcal{D}=\{d_0, d_1, ..., d_m\}$, with spans $\mathcal{S}=\{s_0, s_1, ... s_n\}$ generated by iterating k-grams in each sentence \cite{Wadden2019EntityRA}. Our model aims to jointly solve event mention extraction, event coreference, and entity coreference. 

\textbf{Event Mention Extraction} refers to the subtask of 1) identifying event triggers in $\mathcal{D}$ by predicting the event type for each token $d_i$. 2) Then, given each trigger, corresponding arguments in $\mathcal{S}$ and argument roles are extracted. This task is similar to the sentence-level event extraction task addressed by previous studies \cite{Wadden2019EntityRA, lin-etal-2020-joint}. The difference is that we require extracting \textit{full spans} of all name, nominal, and pronoun arguments, while these works focus on extracting \textit{head spans} of \textit{name} arguments. \textbf{Entity Coreference} aims to find which entity mentions refer to the same entity. Our model predicts the most likely antecedent span $s_j$ for each span $s_i$. \textbf{Event Coreference} is to recognize event mentions that are co-referent to each other. Similar to entity coreference, we predict the most likely antecedent trigger $d_j$ for each  predicted trigger $d_i$. \textbf{Entity Extraction} is performed as an auxiliary subtask for richer representations. Each entity mention corresponds to a span $s_i$ in $\mathcal{S}$.

\subsection{Task Evaluation}
Evaluation metrics used by previous sentence-level event extraction studies \cite{Wadden2019EntityRA, zheng-etal-2019-doc2edag, lin-etal-2020-joint} are not suitable for our task as event coreference and entity coreference are not considered. \citet{du-cardie-2020-document} evaluates entity coreference using bipartite matching. However, it does not consider event coreference and less informative arguments (nominal and pronoun). As a solution, we propose two metrics: \DocTrigger{} and \DocArgument{}, to properly evaluate event extraction at the document level. The purpose is to conduct evaluation on event coreference clusters and argument coreference clusters. \textbf{\DocTrigger{}} considers trigger span, event type, and event coreference. Triggers in the same event coreference chain are clustered together. The metric first aligns gold and predicted trigger clusters, and computes a matching score between each gold-predicted trigger cluster pair. A predicted trigger cluster gets full score if all the associated triggers are correctly identified. To enforce the constraint that one gold trigger cluster can only be mapped to at most one predicted trigger cluster, Kuhn–Munkres algorithm \cite{kuhn1955hungarian} is adopted. \textbf{\DocArgument{}} considers argument span, argument role, and entity coreference. We define an argument cluster as an argument with its co-referent entity mentions. Similar to \DocTrigger{}, \DocArgument{} uses Kuhn–Munkres algorithm to align gold and predicted argument clusters, and compute a matching score between each argument cluster pair. An event extraction system should get full credits in \DocArgument{} as long as it identifies the most informative co-referent entity mentions and does not predict false positive co-referent entity mentions.\footnote{We set the weights for name, nominal, and pronoun to be 1, 0.5, and 0.25, inspired by \citet{chen-ng-2013-linguistically}. } Details of the evaluation metric are included in \Cref{appendix:metrics}.





\section{Proposed Approach}

We develop a base model that makes independent predictions for each subtask under a multi-task IE framework. The proposed end-to-end framework, \DEED~, then incorporates \DVN~ into the base model to efficiently capture cross-event dependencies. 

\begin{figure}[t]
    \centering
    \includegraphics[width=.95\linewidth]{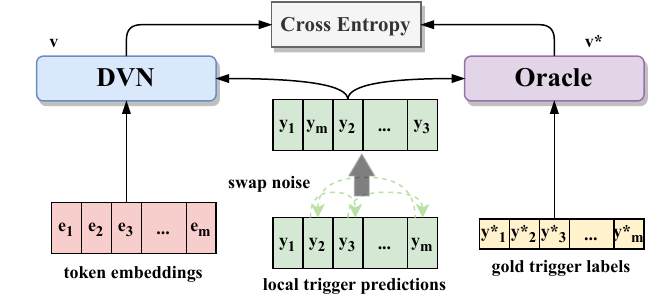}
    \caption{Use swap noise to enable \DVN~ to continue learning from the oracle value function even when the local trigger classifier overfits on the training set.}
    \label{fig:swap_noise}
    \vspace{-5mm}
\end{figure}

\subsection{Base Model}

Our \BASE~ model is built on a span-based IE framework, \DYGIEPP{} \cite{Wadden2019EntityRA}. \DYGIEPP{} learns entity classification, entity corefernce, and event extraction jointly. The base model extends the entity coreference module of \DYGIEPP{} to handle event coreference.
\label{sec:base_model}
\paragraph{Encoding}
Ideally, we want to encode all tokens in a document $\mathcal{D} = \{d_1, d_2, ..., d_m\}$ with embeddings that covers the context of the entire document. However, due to hardware limitation for long documents, each document is split into multi-sentences. Each multi-sentence corresponds to a chunk of consecutive sentences. We obtain rich contextualized embeddings for each multi-sentence of tokens $\bm{e} = \{\bm{e_1}, \bm{e_2}, ..., \bm{e_n}\}$ using \textsc{Bert-Base} \cite{devlin-etal-2019-bert}. 

\paragraph{Span Enumeration}
Conventional event extraction systems use BIO tag scheme to identify the starting and ending position of each trigger and entity. Nevertheless, this method fails to handle nested entities. As a solution, we enumerate all possible spans to generate event mention and entity mention candidates from uni-gram to $k$-gram.\footnote{$k$ is empirically determined to be 12.} Each span $s_i$ is represented by corresponding head token $\bm{e_{h}}$, tail token $\bm{e_{t}}$ and the distance embeddings $\bm{c_{h,t}}$, denoted as $\bm{x_i} = [\bm{e_{h}}, \bm{e_{t}}, \bm{c_{h,t}} ]$, following \citet{Wadden2019EntityRA}.

\paragraph{Classification}
We use task-specific feed-forward networks ($\textrm{FFN}$) to compute the label probabilities. Trigger extraction is performed on each token $\bm{y}^{trig}_i=\textrm{FFN}^{trig}(\bm{e_i})$, while entity extraction is done on each span $\bm{y}^{ent}_i=\textrm{FFN}^{ent}(\bm{x_i})$. For argument extraction, event coreference, and entity coreference, we score each pair of candidate spans $\bm{y^t_{k}} = \textrm{FFN}^t([\bm{x_i},\bm{x_j}])$, where $t$ refers to a specific task. Cross-entropy loss is used to learn trigger extraction, argument extraction as follows
\begin{align*}
    \mathcal{L}^{t} = \frac{1}{N^t}\sum^{N^t}_{i=1} \bm{y^{t*}}_i  \log \bm{y}^t_i,
\end{align*}

, where $\bm{y^{t*}}$ denotes the ground truth labels, $N^t$ denotes the number of instances, and $t$ denotes different tasks.

For entity coreference and event coreference, \BASE~ optimizes marginal log-likelihood for all correct coreferent spans given candidate spans.
\begin{align*}
    \mathcal{L}^{t} = \log \prod^{N}_{i=1} \sum_{j ~\in~ \mathrm{COREF}(i)}\bm{y}^t_{(i,j)},
\end{align*}

where $\mathrm{COREF}(i)$ denotes the gold set of spans coreferent with candidate span $i$, and $t$ denotes different tasks. The total loss function for \BASE~ is the weighted sum of all tasks:
\begin{align*}
    \mathcal{L}^{\BASE~} = \sum_{t} \beta^{t} \mathcal{L}^{t},
\end{align*}
$\beta^t$ is the loss weight for task $t$.

\subsection{Cross-event Dependencies}
\label{section: cross-event dependencies}
A main issue for document-level event extraction is the increased complexity for capturing event dependencies. Due to larger number of events at the scope of document, efficiency is a key challenge to modeling inter-event interactions. We incorporate \DVN~ \cite{Gygli2017DeepVN} into \BASE~ to solve this issue given its advantage in computation efficiency.

\paragraph{Deep Value Networks}
DVN is an energy-based \textit{structured prediction} architecture $v(\bm{x}, \bm{y}; \theta)$ parameterized over $\theta$ that learns to evaluate the compatibility between a structured prediction $\bm{y}$ and an input $\bm{x}$.  The objective of $v(\bm{x}, \bm{y}; \theta)$ is to approximate an \textit{oracle value function} $v^*(\bm{y}, \bm{y}^*)$, a function which measures the quality of the output $\bm{y}$ in comparison to the groundtruth $\bm{y}^*$, $s.t. \forall \bm{y} \in \mathcal{Y}, v(\bm{x}, \bm{y}; \theta) \approx v^*(\bm{y}, \bm{y}^*).$ 
The final evaluation metrics are usually used as the \textit{oracle value function} $v^*(\bm{y}, \bm{y}^*)$. For simplicity, we drop the parameter notion $\theta$ , and use $v(\bm{x}, \bm{y})$ to denote DVN instead.

The inference aims to find $\bm{\hat{y}} = \texttt{argmax}_{\bm{y}}v(\bm{x}, \bm{y})$ for every pair of input and output. A local optimum of $v(\bm{x}, \bm{y})$ can be efficiently found by performing gradient ascent that runs linear in the size of label. Given \DVN~'s higher scalability compared with other structured prediction algorithms, we leverage \DVN~ to capture cross-event dependencies.

\paragraph{Deep Value Networks Integration}
Local trigger classifier predicts the \textit{event type scores} for each token independently. \DVN~ takes in predictions from local trigger classifiers $\bm{y}^{trig}$ and embeddings of all tokens $\bm{e}$ as inputs.
Structured outputs $\bm{\hat{y}}^{trig}$ should correct errors made by the local trigger classifier due to uncaptured cross-event dependencies. $\bm{\hat{y}}^{trig}$ is obtained by performing $h$-iteration updates on local trigger predictions $\bm{y}^{trig}$ using gradient ascent,\footnote{We set $h$=20 for best empirical performance.} 

\begin{align}
    \label{eq:dvn_inference}
    \bm{y}^{t+1} &= \mathcal{P_Y}(\bm{y}^{t} + \alpha \frac{d}{d\bm{y}}v(\bm{e}, \bm{y}^t) ) \nonumber \\
    \bm{\hat{y}}^{trig} &= \bm{y}^{h},
\end{align} 

where $\bm{y}^{1} = \bm{y}^{trig}$, $\alpha$ denotes the inference learning rate, and $\mathcal{P_Y}$ denotes a function that clamps inputs into the range $(0, 1)$. The most likely event type for token $i$ is determined by computing $\texttt{argmax}(\bm{\hat{y}}^{trig}_{i})$.



\paragraph{End-to-end \DVN~ Learning}
We train \DEED~ in an end-to-end fashion by directly feeding the local trigger predictions to both \DVN~ and the oracle value function. The trigger classification $F_1$ metric adopted by previous works \cite{Wadden2019EntityRA, lin-etal-2020-joint} is used as the \textit{oracle value function} $v^*(\bm{y}^{trig}, \bm{y}^{trig*})$. To accommodate continuous outputs, $v^*(\bm{y}^{trig}, \bm{y}^{trig*})$ needs to be relaxed. We relaxed the output label for each token from $[0, 1]$ to $(0, 1)$. Union and intersection set operations for computing the $F_1$ scores are replaced with element-wise minimum and maximum operations, respectively. The relaxed \textit{oracle value function} is denoted as $\underline{v}^*(\bm{y}_{trig}, \bm{y}^*_{trig})$. The loss function for the trigger DVN is the following:

{\small
\begin{align}
    \mathcal{L}^{\DVN~} = &\sum_{\bm{y}^{trig}} - \underline{v}^*(\bm{y}^{trig}, \bm{y}^{trig*})\log v(\bm{e}, \bm{y}^{trig}) \nonumber \\
    & - (1-\underline{v}^*(\bm{y}^{trig}, \bm{y}^{trig*}))\log(1-v(\bm{e}, \bm{y}^{trig})). 
\end{align}}

The total loss function for training \DEED~ end-to-end is the summation of \BASE~ loss and \DVN~ loss,
\begin{align*}
    \mathcal{L}^{\DEED~} = \mathcal{L}^{\BASE~} + \mathcal{L}^{\DVN~}.
\end{align*}
\paragraph{Noise Injection}

However, in this training setup, \DVN~ observes a large portion of high scoring examples at the later stage of training process when the local trigger classifier starts to overfit on the training examples. A naive solution is feeding random noise to train \DVN~ in addition to the outputs of local trigger classifier. Yet, the distribution of these noise are largely distinct from the output of trigger classifier, and therefore easily distinguishable by \DVN~. Thus, we incorporate swap noise into the local trigger predictions, where $s\textrm{\%}$ of the local trigger outputs $\bm{y}^{trig}$ are swapped, as depicted in \Cref{fig:swap_noise}.\footnote{$s$ is empirically set to 20} This way, noisy local trigger predictions have similar distributions to the original trigger predictions. We also hypothesize that higher-confident predictions are often easier to identify, and swapping higher-confident trigger predictions may not help \DVN~ learn. We experimented swapping only the lower-confident trigger predictions.

\begin{table*}[t]
    \centering
    {
    \begin{tabular}{lccccccc}
        \toprule
        & \multicolumn{3}{c}{\textbf{\DocTrigger{}}} & 
        \multicolumn{3}{c}{\textbf{\DocArgument{}}} & \\ \midrule
        Model & Prec. & Rec. &  {F1} &  {Prec.} &  {Rec.} &  {F1} & Comb.   \\ \midrule
        \BASE{} & 71.25 & 60.94 & 65.69 & 43.75 & 48.65 & 46.07 & 17.13 \\
        \DCRF{} & 71.87 & 65.18 & 68.36 & \textbf{49.84} & 52.16 & 50.97 & 34.84 \\ 
        \OneIE{} & 71.96 & 62.04 & 66.63 & 49.64 & \textbf{56.58} & \textbf{52.88} & 35.23 \\
        \cmidrule(lr){1-8}
        \DEED~ & 70.97 & 62.90 & 66.70 &  46.13 & 51.34 & 48.60 & 32.42\\
        
        ~ w/ RN & 71.69 & \textbf{65.76} & 68.59 & 48.52 & 52.53 & 50.44 & 34.60\\
        ~ w/ SN & 70.87 & 64.02 & 67.28 & 43.76 & 55.15 & 48.80 & 32.83\\
        ~ w/ SNLC & \textbf{73.89} & 64.98 & \textbf{69.14} & 48.00 & 55.27 & 51.38 & \textbf{35.52} \\

        \bottomrule
    \end{tabular}
    }
    \vspace{-2mm}
    \caption{Experimental results on ACE05 using document-level evaluation metrics. \textit{RN}: random noise; \textit{SN}: swap noise; \textit{SNLC}: swap noise applying to lower-confident predicted triggers.}
    \label{tab:document_level_result}
\end{table*}
\begin{table*}[t]
    \centering
    {
    \begin{tabular}{@{\ \ }l@{\ \ }c@{\ \ }c@{\ \ }c@{\ \ }c@{\ \ }c@{\ \ }c@{\ \ }c@{\ \ }}
        \toprule
        
        Model & Trig-I & Trig-C & Arg-I &  Arg-C &  Evt-Co &  Ent-Co \\ \midrule
        \DCRF{} & 73.92 & 70.57 & 51.77 & 48.31 & \textbf{54.02} & 74.23 \\ 
        \BASE{} & 71.97 & 68.17 & 47.95 & 44.57 & 43.95 & 71.88   \\ 
        \OneIE{} & 73.91 & 71.01 & \textbf{57.19} & \textbf{53.89} & 42.75 & \textbf{77.00}\\
        \cmidrule(lr){1-7}
        \DEED~  &  73.68 & 69.62 & 52.35 & 48.24 & 53.85 & 75.77\\
        ~ w/ RN & 72.33 & 68.20 & 51.33 & 48.66 & 49.86 & 74.39\\
        ~ w/ SN & 74.19 & 69.54 & 51.27 & 48.10 & 48.94 & 75.60 \\
        ~ w/ SNLC & \textbf{75.06} & \textbf{71.73} & 55.12 & 52.09 & 50.11 & 76.98\\
        
        \bottomrule
    \end{tabular}
    }
    \vspace{-2mm}
    \caption{A breakdown of evaluation for each component in F1 evaluated on ACE05. \textit{Trig}: trigger; \textit{Arg}: argument;\textit{I}: identification; \textit{C}: classification; \textit{Evt-Co}: event coreference; \textit{Ent-Co}: entity coreferecne.}
    \label{tab:sentence_level}
    
\end{table*}

\begin{table}
    \small
    \centering
    {
    
    \begin{tabular}{@{\ \ }l@{\ \ }c@{\ \ }c}
        \toprule
        Model & Training  (sec/ multi-sent) & Inference  (sec/ doc) \\ \midrule
        \BASE{} & 0.52 & 1.50 \\ 
        \DCRF{} & 2.55 & 9.10 \\ 
        \OneIE{} & 1.21 & 15.89\\
        \cmidrule(lr){1-3}
        \DEED{} & 0.71 & 1.52 \\
        \bottomrule
    \end{tabular}
    
    }
    \vspace{-2mm}
    \caption{Comparison of training and inference time, evaluated on the training set and the dev set.}
    \label{tab:training_inference_efficiency}
    \vspace{-5mm}
\end{table}
\section{Experiments}

\subsection{Experimental Setup}

Our models are evaluated on the ACE05 dataset, containing event, relation, entity, and coreference annotations. Experiments are conducted at the document level instead of sentence level as previous works \cite{Wadden2019EntityRA, lin-etal-2020-joint}.

\subsection{Baselines and Model Variations}
We compare \DEED~ with three baselines: (1) \BASE~, the base model described in \Cref{sec:base_model}; (2) \DCRF~ extends \BASE~ by adding a CRF layer on top of the trigger classifier; (3) \OneIE~ is a pipeline composed of the joint model presented in \citet{lin-etal-2020-joint} and coreference modules adapted from \BASE~. \citet{lin-etal-2020-joint} is the state-of-the-art sentence-level event extraction model that utilizes beam search and CRF with global features to model cross sub-task dependencies. For fair comparison, all models are re-trained using \textsc{Bert-Base} \cite{devlin-etal-2019-bert} as the encoder.

In addition to the original \DEED~ model, we consider three variations of it, as discussed in \Cref{section: cross-event dependencies}. \textbf{\DEED~ w/RN} incorporates random noise while learning \DVN~, whereas \textbf{\DEED~ w/SN} integrates swap noise. \textbf{\DEED~ w/SNLC} is an extension of \textbf{\DEED~ w/SN}, where swap noise is only applied to lower-confident trigger predictions.

\subsection{Overall Results}
The overall results are summarized in \Cref{tab:document_level_result}. To measure the overall performance, a combined score (\textit{Comb.}) is computed by multiplying \DocTrigger~ $F_1$ and \DocArgument~ $F_1$. \DEED~ and \DCRF~ achieve huge improvement on all metrics over \BASE~, suggesting the importance of cross-event dependency modeling for our task. Adding random noise or swap noise to train \DVN~ both improve upon the vanilla training method. \OneIE~ achieves the best \DocArgument~ performance, while \textbf{\DEED~ w/SNLC} achieves the highest \DocTrigger~ score and combined score.

\section{Analysis}
\subsection{Performance of Each Component}
To understand the capabilities of each module, we show an evaluation breakdown on each component following previous works \cite{Wadden2019EntityRA, lin-etal-2020-joint} in \Cref{tab:sentence_level}.\footnote{These studies focus on extracting head span of name argument, while we extract full span of all types of arguments.} Both \DCRF~ and \DEED~ obtain significant performance gain over \BASE~ across all tasks. In terms of trigger-related tasks, \textit{Trig-I} and \textit{Trig-C}, \textbf{\DEED~ w/SNLC} achieves the highest scores. Yet, \DCRF~ performs the best on \textit{Evt-Co}. This explains the close performance of \textbf{\DEED~ w/SNLC} and \DCRF~ on \DocTrigger{}, as shown in \Cref{tab:document_level_result}. In terms of argument-related tasks, \OneIE{} achieves the best performance on \textit{Arg-I} and \textit{Arg-C}. This suggests that cross-subtask modeling can be important to improve argument extraction. \textit{Arg-I} and \textit{Arg-C} are much lower than the reported scores by previous studies ~\cite{Wadden2019EntityRA, lin-etal-2020-joint}. This suggests the difficulty of extracting full span of pronoun and nominal arguments. 


\subsection{Computation Time} \Cref{tab:training_inference_efficiency} describes the computation time of different models. \DEED~ only requires slightly more computation time in both training and inference time than \BASE~. By contrast, compared to \DCRF{}, \DEED~ is $\sim$3.5x faster in training time and $\sim$6x faster in inference time. This demonstrates the efficiency of our approach given the little increase in computation time and the significant performance gain comparable to \DCRF~ detailed in \Cref{tab:document_level_result,tab:sentence_level}. We also added experiments with \OneIE{} as a reference, but the comparison focuses on end-to-end frameworks.

\subsection{Value Function Approximation}
To show that the performance gain of \DEED~ is resulted from improved capabilities of \DVN~ in judging the structure of predicted triggers, we investigate how close \DVN~ approximates the oracle value function under different training settings. We use cross entropy loss as the distance function between the output of \DVN~ and and output of the oracle value function on the test set. The lower the loss is, the closer between the output of \DVN~ and the output of the oracle value function. \Cref{tab:dvn_quality} shows the approximation results. The SNLC variation (swap noise applying to lower-confident predicted triggers) yields the lowest loss comparing to the base model and other variations. Along with the results shown in \Cref{tab:sentence_level}, we show that lower \DVN~ loss results in better trigger scores. This demonstrates that integrating noise into \DVN~ training procedure is effective in learning better \DVN~ and obtaining better overall performance.

\begin{table}[t]
    \small
    \centering
    {
    \begin{tabular}{lc}
        \toprule
        
        Training Method & Loss (Cross Entropy)   \\ \midrule

        Original & 0.3613   \\
        RN &  0.7451 \\
        SN &  0.2393 \\
        SNLC & 0.2298 \\

        \bottomrule
    \end{tabular}
    }
    \vspace{-2mm}
    \caption{The average \DVN~ loss of different \DEED~ training methods on the test set. The lower the loss, the closer between \DVN~ and the \textit{oracle value function}.}
    \label{tab:dvn_quality}
    \vspace{-5mm}
\end{table}

\subsection{Error Analysis} 
\begin{figure}[h]
    \centering
    \includegraphics[width=.95\linewidth]{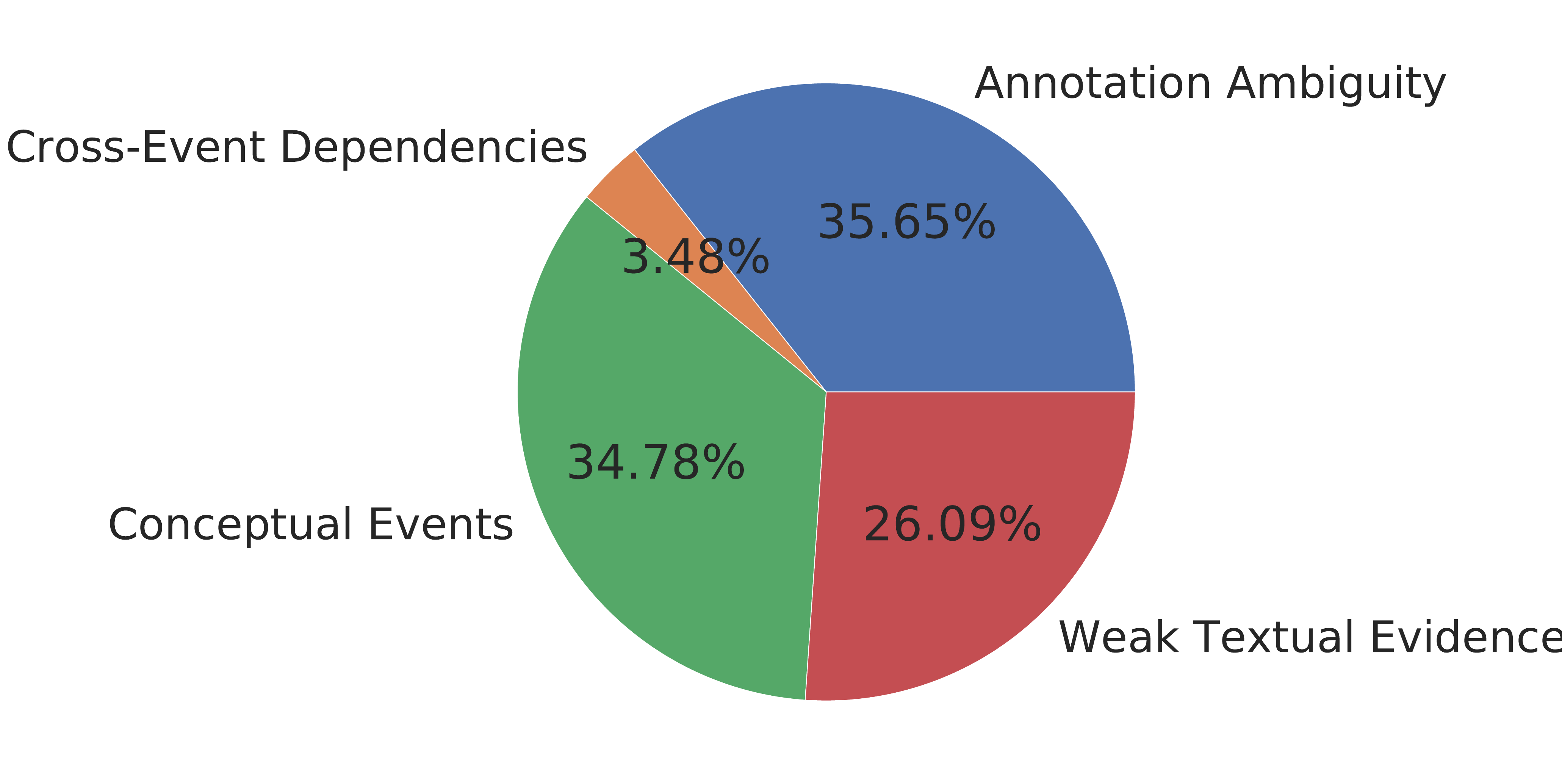}
    \caption{Distribution of errors made by \DVN~ on the ACE05 test set.}
    
    \label{fig:error_distribution}
\end{figure}

We manually compared gold and predicted labels of event mentions on the ACE05 test set and analyzed the mistakes made by our model. These errors are categorized as demonstrated in \Cref{fig:error_distribution}.

\paragraph{Annotation ambiguity}
A significant portion of the false positive errors are caused by the ambiguity of the task. Such ambiguity can result in disagreement between human annotators. For example,

\begin{quotation}
  \small
  Lebanese Prime Minister Rafiq Hariri submitted his \textit{resignation} Tuesday and it was accepted by President Emile Lahoud.
  
\end{quotation}
In the sentence above, the trigger label for token \textit{resignation} should be \textsc{End-Position}, according to the annotation guideline. Yet, it is not annotated as a trigger in gold annotation. In other cases, two sentences with similar structures contain inconsistent gold annotation, such as:
\begin{quotation}
  \small
  Separately, \textit{former} WorldCom CEO Bernard Ebbers failed on April 29 to make a first repayment of 25 million dollars ...
\end{quotation}

\begin{quotation}
  \small
  \textit{Former} senior banker Callum McCarthy begins what is one of the most important jobs in London 's financial world in September 
\end{quotation}
The two examples above share similar context. However, the \textit{former} in the first sentence is not involved with any event, whereas the \textit{former} in the second sentence is annotated as an \textsc{End-Position} typed trigger.

\paragraph{Conceptual Events}
Another common source of false positive errors is extracting ``conceptual" events, which did not happen or may happen in the future. For instance,
\begin{quotation}
    \small
    ... former WorldCom CEO Bernard Ebbers failed on April 29 to make a first \textit{repayment} of 25 million dollars ...
\end{quotation}
Our model predicts the word \textit{repayment} as an \textsc{Transfer-Money}, which is true if it indeed happened, except it \textit{failed}, as indicated in the beginning of the sentence. To handle this type of error, models need to be aware of the tense and whether there is a negative sentiment associated with the predicted events.

\paragraph{Weak Textual Evidence}
Our model commonly made false negative errors in cases where the textual information is vague.

\begin{quotation}
    \small
    But both men observed an uneasy truce over US concerns about Russian \textit{aid} to the nuclear program of Iran ...
\end{quotation}
In the above sentence, \DVN~ fails to identify the token \textit{aid} as a trigger of type \textsc{Transfer-Money}. In fact, it is hard to determine whether the \textit{aid} is monetary or military given the context of the whole document. In this case, models have to be aware of information from other sources, such as knowledge bases or other news articles.

\paragraph{Cross-event Dependencies}
Although our model is able to correct many mistakes made by \BASE~ that requires modeling of cross-event dependencies, as demonstrated in \Cref{tab:cross_sentence_comparison}, there are still a few cases where our model fails.
\begin{quotation}
    \small
    ... after the city 's bishop committed \textit{suicide} over the 1985 blasphemy law . Faisalabad 's Catholic Bishop John Joseph , who had been campaigning against the law , \textit{shot} himself in the head outside a court in Sahiwal district when the judge ... himself in the head outside a court
\end{quotation}
In the above example, \DVN~ correctly predict \textit{suicide} as a \textsc{Die} typed trigger, but falsely predict \textit{shot} as type \textsc{Attack} instead of type \textsc{Die}. If our model could capture the interactions between \textit{suicide} and \textit{shot}, it would be able to process this situation. There is still room to improve in cross-event dependency modeling.

\begin{table}[t]
    \centering
    {
    \small
    \begin{tabular}{cc|cc}
        \toprule
         & & Within sentence & Cross sentence \\ \midrule
        \multirow{2}{*}{\BASE~} & Correct & 161 & 126  \\
        \cmidrule(lr){2-4} 
        & Incorrect & 71 & 45 \\
        \cmidrule(lr){1-4}
        \multirow{2}{*}{\DEED~} & Correct & 166 & 136  \\
        \cmidrule(lr){2-4} 
        & Incorrect & 66 & 35 \\
        \bottomrule 
    \end{tabular}
    }
    \vspace{-2mm}
    \caption{Trigger predictions comparison between \BASE~ and \DEED~. \textit{Cross sentence} refers to triggers with co-referent triggers that lie in different sentences.}
    \label{tab:cross_sentence_comparison}
    \vspace{-5mm}
\end{table}

\section{Conclusion}
In this paper, we investigate document-level event extraction that requires joint modeling of event and entity coreference. We propose a document-level event extraction framework, \DEED~, which uses \DVN~ to capture cross-event dependencies, and explore different end-to-end learning methods of \DVN~. Experimental results show that \DEED~ achieves comparable performance to competitive baseline models, while \DEED~ is much favorable in terms of computation efficiency. We also found that incorporating noise into end-to-end \DVN~ training procedure can result in higher \DVN~ quality and better overall performance. 

\section{Ethics}
Biases have been studied in many information extraction tasks, such as relation extraction \cite{gaut-etal-2020-towards}, named entity recognition \cite{10.1145/3372923.3404804}, and coreference resolution \cite{zhao-etal-2018-gender}. Nevertheless, not many works investigate biases in event extraction tasks, particularly ACE05. 

We analyze the portion of male pronouns (he, him, and his) and female pronouns (she and her) in the ACE05 dataset. In total, there are 2780 male pronouns, while only 970 female pronouns appear in the corpus. We would expect the trained model to perform better when extracting events where male arguments are involved, and make more mistakes for event involving female arguments due to the significant imbalance between male and female entity annotation. After analyzing the performance of \textbf{\DEED~ w/ SNLC} on the test set, we found that it scores 54.90 and 73.80 on \textit{Arg-C} $F_1$ for male and female pronoun arguments, respectively. Surprisingly, our model is better at identifying female pronoun arguments than male pronoun arguments. 

While our proposed framework may not subject to gender biases in ACE05, whether such issue can occur when our model is deployed for public use is unknown. Rigorous studies on out-of-domain corpus is needed to answer this question.

\section*{Acknowledgements}
We appreciate insightful feedback from PLUSLab members and the anonymous reviewers. This research was sponsored by the Intelligence Advanced Research Projects Activity (IARPA), via Contract No. 2019-19051600007. The views and conclusions of this paper are those of the authors and do not reflect the official policy or position of IARPA or the US government.
\bibliographystyle{acl_natbib}
\typeout{}
\bibliography{naacl2021}

\clearpage
\appendix



    

\section{Data Statistics}
The statistics of ACE05 are shown in \Cref{tab:data_stats}.We observe that the event coreference annotation is very sparse. 
\begin{table}[h]
    \centering
    {
    \small
    \begin{tabular}{cccccc}
        \toprule
        Split & Docs & Events & Entities & Ent-C & Evt-C \\ \midrule
        Train & 529 & 4202 & 47569 & 6814 & 482 \\
        Dev & 28 & 450 & 3423 & 553  & 45 \\ 
        Test & 40 & 403 & 3673 & 577  & 58 \\
        \bottomrule 
    \end{tabular}
    }
    \vspace{-2mm}
    \caption{Data statistics of ACE05. \textit{Ent-C} and \textit{Evt-C} denote the number of entity and event coreference clusters, respectively.}
    \label{tab:data_stats}
    \vspace{-5mm}
\end{table}
\section{Implementation Details}
We adopted part of the pre-processing pipelines from \citet{Wadden2019EntityRA} for data cleaning and dataset splitting.

\BASE{}, \DCRF{}, and \DVN~ are optimized with \textsc{BertAdam} for 250 epochs with batch size of 16. \textsc{Bert-Base} is fine-tuned with learning rate of 1e-4 and no decay, while the other components are trained with learning rate of 1e-3 and weight decay of 1e-2. Training is stopped if the dev set \textit{Arg-C} $F_1$ score does not improve for 15 consecutive epochs. \OneIE~ is trained with the default parameters described in \citet{lin-etal-2020-joint}. All experiments are conducted on a 12-CPU machine running CentOS Linux 7 (Core) and NVIDIA RTX 2080 with CUDA 10.1.

\section{Document-level Evaluation Metrics}
\label{appendix:metrics}
\begin{algorithm}
    \caption{Document-level Trigger Evaluation Metric}
    \begin{algorithmic}[1]
      \Function{\DocTrigger~}{gold events $G$, predicted events $P$} 
            \State Let \textit{match} = \textit{false-alarm} = \textit{miss} = \textit{hit} = 0
            \State Let $M$ be a trigger matching matrix.
           \For { $g$ in $G$.triggers}
                \For {$p$ in $P$.triggers}
                    
                    \If {! \textsc{SameEventType}($g$, $p$)}
                        \State \textit{match} = 0
                    \Else
                        \State \textit{match} = Trig-I($p$, $g$) 
                    \EndIf
                    \State  $M$[\textit{g.idx}, \textit{p.idx}] = \textit{match}
                    
                \EndFor
           \EndFor
           \State \textit{assignments} = \textsc{Kuhn-Munkres}($M$)
            
            \For {$i,j$ in \textit{assignments}}
                \If  {$G$.triggers[$i$] is null}
                    \State \textit{false-alarm} += 1
                \ElsIf {$P$.triggers[$j$] is null}
                    \State \textit{miss} += 1
                \Else
                    \State \textit{match} += $M$[$i$][$j$]
                    \State \textit{hit} += 1
                \EndIf
            \EndFor
        \State \Return (\textit{match}, \textit{false-alarm}, \textit{miss}, \textit{hit})
       \EndFunction
    
\end{algorithmic}

\end{algorithm}

\begin{algorithm}
    
    \caption{Document-level Argument Evaluation Metric}
    \begin{algorithmic}[1]
      \Function{\DocArgument~}{gold events $G$, predicted events $P$} 
            \State Let \textit{match} = \textit{false-alarm} = \textit{miss} = \textit{hit} = 0
            \State Let $M$ be an argument matching matrix.
           \For {$g$ in $G$.arguments}
                \For{$p$ in $P$.arguments}
                    
                    \State  $M$[$i$, $j$] = \textsc{ArgMatch}($g$, $p$)
                \EndFor
           \EndFor
           \State \textit{assignments} = \textsc{Kuhn-Munkres}($M$)
            
            \For {$i,j$ in \textit{assignments}}
                \If  {$G$.arguments[$i$] is null}
                    \State \textit{false-alarm} += 1
                \ElsIf {$P$.arguments[$j$] is null}
                    \State \textit{miss} += 1
                \Else
                    \State \textit{match} += $M$[$i$][$j$]
                    \State \textit{hit} += 1
                \EndIf
            \EndFor
        \State \Return (\textit{match}, \textit{false-alarm}, \textit{miss}, \textit{hit})
       \EndFunction
    
\end{algorithmic}
\label{arg:docargument}
\end{algorithm}

\begin{algorithm}
    
    \caption{Argument match called by \Cref{arg:docargument}}
    \begin{algorithmic}[1]
    \Function{ArgMatch}{gold argument cluster $g$, predicted argument cluster $p$} 
        \If{ not \textsc{SameRole}($g$, $p$) or not 
        \State not \textsc{SameEventType}($g$, $p$)} 
            \State \Return $0$
        \EndIf
        \State \textit{BMA} = \textsc{BestMatchedArgument}($p$,$g$) 
        \State \textit{w} = \textsc{GetWeight}(\textit{BMA}) \Comment{The weights for name, nominal, pronoun are 1, 0.5, 0.25. }
        \State \textit{false-alarm} = $|p - g|$ \Comment{Set operation}
        \State \Return $w \times (1 - \frac{\textit{false-alarm}}{|p|})$
        
    \EndFunction
    
\end{algorithmic}

\end{algorithm}

\section{Development Set Performance}
\begin{table*}[t]
    \centering
    {
    \begin{tabular}{lcccccc}
        \toprule
        
        Model & Trig-I & Trig-C & Arg-I &  Arg-C & Evt-Co &  Ent-Co   \\ \midrule
        \BASE{} &  74.63 & 70.49 & 56.82 & 52.41  & 30.64 & 67.31  \\ 
        \DCRF{} & 76.53 & 72.89 & 59.62 & 54.47 & 33.16 & 68.72 \\ 
        \OneIE{} & 76.78 & 73.56 & \textbf{63.12} & \textbf{59.32} & 35.81 & \textbf{70.78}   \\ 
        \cmidrule(lr){1-7}
        \DEED~  &  77.11 & 72.31 & 62.42 & 55.80 & 31.90 & 69.57\\
        ~~ w/ RN & 75.74 & 70.94 & 61.45 & 55.18  & 34.88 & 68.56\\
        ~~ w/ SN & \textbf{77.81} & \textbf{74.53} & 61.90 & 55.52 & \textbf{38.55} & 69.48 \\
        ~~ w/ SNLC & 76.76 & 72.13 & 62.78 & 57.45 & 31.32 & \textbf{70.78}\\
        
        \bottomrule
    \end{tabular}
    }
    \vspace{-2mm}
    \caption{A breakdown of evaluation on the dev set for each model. The corresponding test set performance is shown in \Cref{tab:sentence_level}.}
    \label{tab:dev_sentence_level}
    \vspace{-5mm}
    
\end{table*}

\end{document}